\documentclass{article}

\usepackage{times}
\usepackage{graphicx} 
\usepackage{subfigure}

\usepackage{amsmath}

\usepackage{natbib}

\usepackage{algorithm}
\usepackage{algorithmic}

\usepackage{hyperref}
\usepackage{url}

\usepackage{booktabs}
\usepackage{color}

\usepackage[accepted]{icml2017}

\begin{document} 

\twocolumn[\icmltitle{A Joint Model for Question Answering and Question Generation}



\icmlsetsymbol{equal}{*}
\begin{icmlauthorlist}
\icmlauthor{Tong Wang}{equal,malu}
\icmlauthor{Xingdi Yuan}{equal,malu}
\icmlauthor{Adam Trischler}{malu}
\end{icmlauthorlist}
\icmlaffiliation{malu}{Microsoft Maluuba}
%
\icmlcorrespondingauthor{Tong Wang}{tong.wang@microsoft.com}
%

\vskip 0.3in ]



\printAffiliationsAndNotice{\icmlEqualContribution} 

\begin{abstract} 
  We propose a generative machine comprehension model that learns 
  jointly to ask and answer questions based on documents. The proposed model
  uses a sequence-to-sequence framework that encodes the document and generates
  a question (answer) given an answer (question). Significant improvement in
  model performance is observed empirically on the SQuAD corpus, confirming our hypothesis that the
  model benefits from jointly learning to perform both tasks. We believe the
  joint model's novelty offers a new perspective on machine comprehension
  beyond architectural engineering, and serves as a first step towards
  autonomous information seeking.
\end{abstract}

\section{Introduction}
\label{sec:intro}
Question answering (QA) is the task of automatically producing an answer to a question
given a corresponding document. It not only provides humans with efficient access to vast
amounts of information, but also acts as an important proxy task to assess machine
literacy via reading comprehension. Thanks to the recent release of several
large-scale machine comprehension/QA datasets
\citep{hermann2015teaching,rajpurkar2016squad,dunn2017searchqa,trischler2016newsqa,nguyen2016ms},
the field has undergone significant advancement, with an array of neural
models rapidly approaching human parity on some of these benchmarks \cite{rnet,shen2016reasonet,seo2016bidirectional}.
However, previous models do not treat QA as a task of natural language generation (NLG),
but of pointing to an answer span within a document.

Alongside QA, question generation has also gained increased popularity~\citep{du2017learning,yuan2017machine}. The task
is to generate a natural-language question conditioned on an answer and the
corresponding document. Among its many applications, question generation has
been used to improve QA systems
\citep{buck2017ask,serban2016generating,yang2017semi}.
A recurring theme among previous studies is to augment existing labeled data with
machine-generated questions; to our knowledge, the direct (though implicit) effect of
asking questions on answering questions has not yet been explored. 

In this work, we
propose a joint model that both asks and answers questions, and
investigate how this joint-training setup affects the individual tasks.
We hypothesize that question generation can help models achieve better QA
performance. This is motivated partly by observations made in
psychology that devising questions while reading can increase
scores on comprehension tests \citep{singer1982active}.
Our joint model also serves as a novel framework for improving QA performance
outside of the network-architectural engineering that characterizes
most previous studies.

Although the question answering and asking tasks appear symmetric, there are some key differences.
First, answering the questions in most existing QA datasets is \emph{extractive} --- it requires selecting some span of text within the document --- while question asking is comparatively \emph{abstractive} --- it requires generation of text that may not appear in the document.
Furthermore, a (document, question) pair typically specifies a unique answer. Conversely, a typical (document, answer) pair may be associated with multiple questions, since a valid question can be formed from any information or relations which uniquely specify the given answer.

To tackle the joint task, we construct an attention-based
\citep{DBLP:journals/corr/BahdanauCB14} sequence-to-sequence model
\citep{sutskever2014sequence} that takes a document as input and generates a
question (answer) conditioned on an answer (question) as output. To address the mixed extractive/abstractive
nature of the generative targets, we use the
pointer-softmax mechanism \citep{gulcehre2016pointing} that learns to
switch between copying words from the document and
generating words from a prescribed vocabulary. Joint training is realized
by alternating the input data between question-answering and question-generating
examples for the same model. We demonstrate empirically that this model's QA performance on \emph{SQuAD},
while not state of the art, improves by about 10\% with joint training.
A key novelty of our joint model is that it can generate (partially) abstractive answers.

\section{Related Work}
\label{sec:related-work}
Joint-learning on multiple related tasks has been explored previously
\citep{collobert2011natural,firat2016multi}.
In machine translation, for instance, \citet{firat2016multi}
demonstrated that translation quality clearly
improves over models trained with a single language pair when the attention mechanism in a neural translation model is shared
and jointly trained on multiple language pairs.


In question answering, \citet{wang2016machine} proposed one of the first neural models for the
SQuAD dataset. SQuAD defines an \emph{extractive} QA task wherein answers consist
of word spans in the corresponding document.
\citet{wang2016machine} demonstrated that
learning to point to answer boundaries is more effective than learning to
point sequentially to the tokens making up an answer span.
Many later studies adopted this boundary model and achieved
near-human performance on the task
\citep{rnet,shen2016reasonet,seo2016bidirectional}.
However, the boundary-pointing mechanism is not
suitable for more open-ended tasks, including abstractive QA \citep{nguyen2016ms}
and question generation. While ``forcing'' the extractive boundary model onto abstractive
datasets currently yields state-of-the-art results
\citep{rnet}, this is mainly because current generative models are poor
and NLG evaluation is unsolved.

Earlier work on question generation has resorted to either rule-based reordering methods
\citep{heilman2010good,agarwal2011automatic,ali2010automation} or slot-filling
with question templates
\citep{lindberg2013generating,chali2016ranking,labutov2015deep}. These
techniques often involve pipelines of independent components that are difficult
to tune for final performance measures.
Partly to address this limitation, end-to-end-trainable neural models have recently been
proposed for question generation in both vision
\citep{mostafazadeh2016generating} and language. For example, \citet{du2017learning}
used a sequence-to-sequence model with an attention mechanism derived
from the encoder states. \citet{yuan2017machine} proposed a similar architecture
but in addition improved model performance through policy gradient techniques.

Several neural models with a questioning component have been proposed for the
purpose of improving QA models, an objective shared by this study. \citet{yang2017semi} devised a semi-supervised training framework that
trained a QA model \citep{dhingra2016gated} on both labeled data and artificial
data generated by a \emph{separate} generative component. \citet{buck2017ask}
used policy gradient with a QA reward to train a sequence-to-sequence paraphrase
model to reformulate questions in an existing QA dataset
\citep{dunn2017searchqa}. The generated questions were then used to further train an
existing QA model \citep{seo2016bidirectional}. A key distinction of our model
is that we harness the \emph{process} of asking questions to 
benefit question answering, without training the model to answer the
generated questions.

\section{Model Description}
\label{sec:model}
Our proposed model adopts a sequence-to-sequence framework
\citep{sutskever2014sequence} with an attention mechanism
\citep{DBLP:journals/corr/BahdanauCB14} and a pointer-softmax decoder
\citep{gulcehre2016pointing}. Specifically, the model takes a document
(i.e., a word sequence) $D = (w^d_1,\dots,w^d_{n_d})$ and a condition sequence
$C = (w^c_1,\dots,w^c_{n_c})$ as input, and outputs a target sequence $Y^{\{q,a\}} = (\hat
w_1,\dots,\hat w_{n_p})$. The condition corresponds to the question word
sequence in answer-generation mode (a-gen), and the answer word sequence in question-generation mode (q-gen).
We also attach a binary variable to indicate
whether a data-point is intended for a-gen or q-gen. Intuitively, this should help the model
learn the two modalities more easily. Empirically, QA performance improves slightly with this addition.

\subsection*{Encoder}
A word $w_i$ in an input sequence is first embedded with an embedding layer into vector
${\bf e}^w_i$. Character-level information is captured
with the final states ${\bf e}^{ch}_i$ of a bidirectional Long Short-Term Memory model
\citep{hochreiter1997long} on the character sequences of $w_i$. The final
representation for a word token ${\bf e}_i=\langle{\bf e}^w_i,{\bf e}^{ch}_i\rangle$ concatenates
the word- and character-level embeddings. These are subsequently encoded with
another BiLSTM into annotation vectors ${\bf h}^d_i$ and ${\bf h}^c_j$ (for the
document and the condition sequence, respectively).

To better encode the condition, we also extract the encodings of the document
words that appear in the condition sequence. This procedure is particularly helpful in
q-gen mode, where the condition (answer) sequence is typically
extractive. These extracted vectors are then fed
into a condition aggregation BiLSTM to produce the \emph{extractive condition encoding} ${\bf h}^e_k$.
We specifically take the final states of the condition encodings
${\bf h}^c_J$ and ${\bf h}^e_K$.
To account for the different extractive vs. abstractive nature of questions vs. answers, we use ${\bf h}^c_J$ in a-gen mode (for encoding questions) and ${\bf h}^e_K$ in q-gen mode (for encoding answers).

\subsection*{Decoder}
The RNN-based decoder employs the pointer-softmax mechanism
\citep{gulcehre2016pointing}. At each generation
step, the decoder decides adaptively whether (a) to generate from a decoder vocabulary or (b) to point to a word in
the source sequence (and copy over). Recurrence of the pointing decoder is
implemented with two LSTM cells $c_1$ and $c_2$:
\begin{eqnarray}
 \label{eqn:dec_lstm1}
 {\bf s}_1^{(t)} & = & c_1({\bf y}^{(t-1)}, {\bf s}_2^{(t-1)})\\
 \label{eqn:dec_lstm2}
 {\bf s}_2^{(t)} & = & c_2({\bf v}^{(t)}, {\bf s}_1^{(t)}),
\end{eqnarray}
where ${\bf s}_1^{(t)}$ and ${\bf s}_2^{(t)}$ are the recurrent states, ${\bf y}^{(t-1)}$
is the embedding of decoder output from the previous time step,
and ${\bf v}^{(t)}$ is the context vector (to be defined shortly in
Equation~\eqref{eqn:wde}).

The pointing decoder computes a distribution $\boldsymbol{\alpha}^{(t)}$ over the
document word positions (i.e., a document attention,
\citealt{DBLP:journals/corr/BahdanauCB14}). Each element is defined as:
\begin{eqnarray*}
 \alpha^{(t)}_i = f({\bf h}^d_i, {\bf h}^c, {\bf h}^e, {\bf s_1}^{(t-1)}),
\end{eqnarray*}
where $f$ is a two-layer MLP with \emph{tanh} and \emph{softmax} activation, respectively.
The context vector ${\bf v}^{(t)}$ used in Equation~\eqref{eqn:dec_lstm2} is the sum of the document encoding weighted by the
document attention:
\begin{eqnarray}
  {\bf v}^{(t)}=\sum_{i=1}^n \alpha^{(t)}_i{\bf h}^d_i.
  \label{eqn:wde}
\end{eqnarray}

The generative decoder, on the other hand, defines a distribution over a
prescribed decoder vocabulary with a two-layer MLP $g$:
\begin{eqnarray}
  {\bf o}^{(t)}=g({\bf y}^{(t-1)},{\bf s}_2^{(t)},{\bf v}^{(t)},{\bf h}^c,{\bf h}^e).
  \label{eqn:smx_gen}
\end{eqnarray}
Finally, the switch scalar $s^{(t)}$ at each time step is computed by a
three-layer MLP $h$:
\begin{eqnarray*}
  s^{(t)}=h({\bf s}_2^{(t)},{\bf v}^{(t)},\boldsymbol{\alpha}^{(t)},{\bf o}^{(t)}),
\end{eqnarray*}
The first two layers of $h$ use \emph{tanh} activation and the final layer uses
\emph{sigmoid} activation, and highway connections are present between the first
and the second layer. We also attach the entropy of the softmax distributions to the input of the final layer, postulating
that the quantities should help guide the switching  mechanism by indicating the confidence of pointing vs generating. The addition is empirically observed to improve model performance.
  
The resulting switch is used to interpolate the pointing
and the generative probabilities for predicting the next word:
\begin{eqnarray*}
  p(\hat w_t)\sim s^{(t)} \boldsymbol{\alpha}^{(t)} + (1-s^{(t)}){\bf o}^{(t)}.
\end{eqnarray*}

\section{Training and Inference}
\label{sec:train}
The optimization objective for updating the model parameters $\theta$ is to maximize the
negative log likelihood of the generated sequences with respect to the training
data $\mathcal{D}$:
\begin{eqnarray*}
  \mathcal{L}=-\sum_{x\in \mathcal{D}}\log p(\hat w_t|w_{<t},x;\theta).
\end{eqnarray*}
Here, $w_{<t}$ corresponds to the embeddings ${\bf y}^{(t-1)}$ in
Equation~\eqref{eqn:dec_lstm1} and~\eqref{eqn:smx_gen}. During training, gold targets
are used to teacher-force the sequence generation for training, i.e.,
$w_{<t}=w^{\{q,a\}}_{<t}$, while during inference, generation is conditioned on
the previously generated words, i.e., $w_{<t}=\hat w_{<t}$.

For words with multiple occurrence, since their exact references in the document cannot be reiabled determined, we aggregate the
probability of these words in the encoder and the pointing decoder (similar to
\citealt{kadlec2016text}). At test time, beam search is used to enhance fluency in
the question-generation output.\footnote{The effectiveness of beam search can be undermined by the generally diminished output length. We therefore do not use beam search in a-gen mode, which also saves training time.} The decoder also keeps an explicit history of previously generated
words to avoid repetition in the output.

\section{Experiments}
\label{sec:experiments}
\subsection{Dataset}
\label{sec:dataset}
We conduct our experiments on the SQuAD corpus \citep{rajpurkar2016squad}, a machine
comprehension dataset consisting of over 100k crowd-sourced question-answer
pairs on 536 Wikipedia articles. Simple preprocessing is performed, including
lower-casing all texts in the dataset and using \emph{NLTK} \citep{bird2006nltk}
for word tokenization. The test split of SQuAD is hidden from the public. We therefore 
take 5,158 question-answer pairs (self-contained in 23 Wikipedia articles) from the training set 
as validation set, and use the official development data to
report test results. Note that answers in this dataset are strictly extractive, and we therefore constrain the pointer-softmax module to point
at all decoding steps in answer generation mode.

\subsection{Baseline Models}
We first establish two baselines without multi-task training. Specifically,
model \texttt{A-gen} is trained only to generate an answer given a document and
a question, i.e., as a conventional QA model. Analogously, model
\texttt{Q-gen} is trained only to generate questions from documents and answers.
Joint-training (in model \texttt{JointQA}) is realized by feeding answer-generation and
question-generation data to the model in an alternating fashion between mini-batches.

In addition, we compare answer-generation performance with the \emph{sequence model}
variant of the match-LSTM (\texttt{mLSTM}) model \citep{wang2016machine}. As mentioned
earlier, in contrast to existing neural QA models that point to the start
and end boundaries of extractive answers, this model predicts a sequence of
document positions as the answer. This makes it most comparable to our QA setup.
Note, however, that our model has the additional capacity to generate abstractively from the decoder vocabulary.

\subsection{Quantitative Evaluation}
We use \emph{F1} and \emph{Exact Match}
(\emph{EM}, \citealt{rajpurkar2016squad}) against the gold answer sequences to
evaluate answer generation, and \emph{BLEU}\footnote{We use the \emph{Microsoft COCO Caption Evaluation} scripts (\url{https://github.com/tylin/coco-caption}) to calculate \emph{BLEU} scores.} \citep{papineni2002bleu} against the
gold question sequences to evaluate question generation. However, existing
studies have shown that the task of question generation often exhibits
linguistic variance that is semantically admissible; this renders it
inappropriate to judge a generated question solely by matching against a gold sequence
\citep{yuan2017machine}. We therefore opt to assess the quality of generated
questions $Y^q$ with two pretrained neural models as well: we use a language model to compute the perplexity of $Y^q$, 
and a QA model to answer $Y^q$. We measure the \emph{F1} score of the answer produced by this QA model.

We choose \texttt{mLSTM} as the pretrained QA model and train it on SQuAD with the same split as mentioned in Section~\ref{sec:dataset}. Performance on the test set (i.e., the official validation set of SQuAD) is 73.78 \emph{F1} and 62.7 \emph{EM}. For the pretrained language model, we train a single-layer LSTM language model on the combination of the \emph{text8} corpus\footnote{\url{http://mattmahoney.net/dc/textdata}}, the \emph{Quora Question Pairs} corpus\footnote{\url{https://data.quora.com/First-Quora-Dataset-Release-Question-Pairs}}, and the gold questions from SQuAD. The latter two corpora were included to tailor to our purpose of assessing \emph{question} fluency, and for this reason, we ignore the semantic equivalence labels in the Quora dataset. Validation perplexity is 67.2 for the pretrained language model.

\subsection{Analysis and Discussion}
\begin{table}[t!]
  \caption{Model evaluation on question- and answer-generation. }
  \label{tab:da_table}
  \resizebox{.48\textwidth}{!}{
    \begin{tabular}{r|cc|ccc}
      \toprule
      & \multicolumn{2}{c|}{\it Answer Generation} & \multicolumn{3}{c}{\it Question
        Generation}\\
      Models & F1 & EM & QA$_\text{F1}$ & PPL & BLEU$_4$\\
      \midrule
      \texttt{A-gen} & 54.5 & 41.0 & \multicolumn{3}{c}{--} \\
      \texttt{Q-gen} & \multicolumn{2}{c|}{--} & 72.4 & 260.7 & 10.8\\
      \texttt{JointQA} & 63.8 & 51.7 & 71.6 & 262.5 & 10.2\\
      \midrule
      \texttt{mLSTM} & 68.2 & 54.4 & \multicolumn{3}{c}{--} \\
      \bottomrule
    \end{tabular}
  }
\end{table}

Evaluation results are provided in Table~\ref{tab:da_table}. We see that A-gen performance improves significantly with the joint model: both \emph{F1} and \emph{EM} increase by about 10 percentage points. Performance of q-gen worsens after joint training, but the decrease is relatively small. Furthermore, as pointed out by earlier studies, automatic metrics often do not correlate well with the generation quality assessed by humans \citep{yuan2017machine}. We thus consider the overall outcome to be positive.

Meanwhile, although our model does not perform as well as \texttt{mLSTM} on the QA task, it has the added capability of generating questions.
\texttt{mLSTM} uses a more advanced encoder tailored to QA, while our model uses only a bidirectional LSTM for encoding.
Our model uses a more advanced decoder based on the pointer-softmax that enables it to generate abstactively and extractively.

\begin{figure}
    \centering
    \includegraphics[width=.48\textwidth]{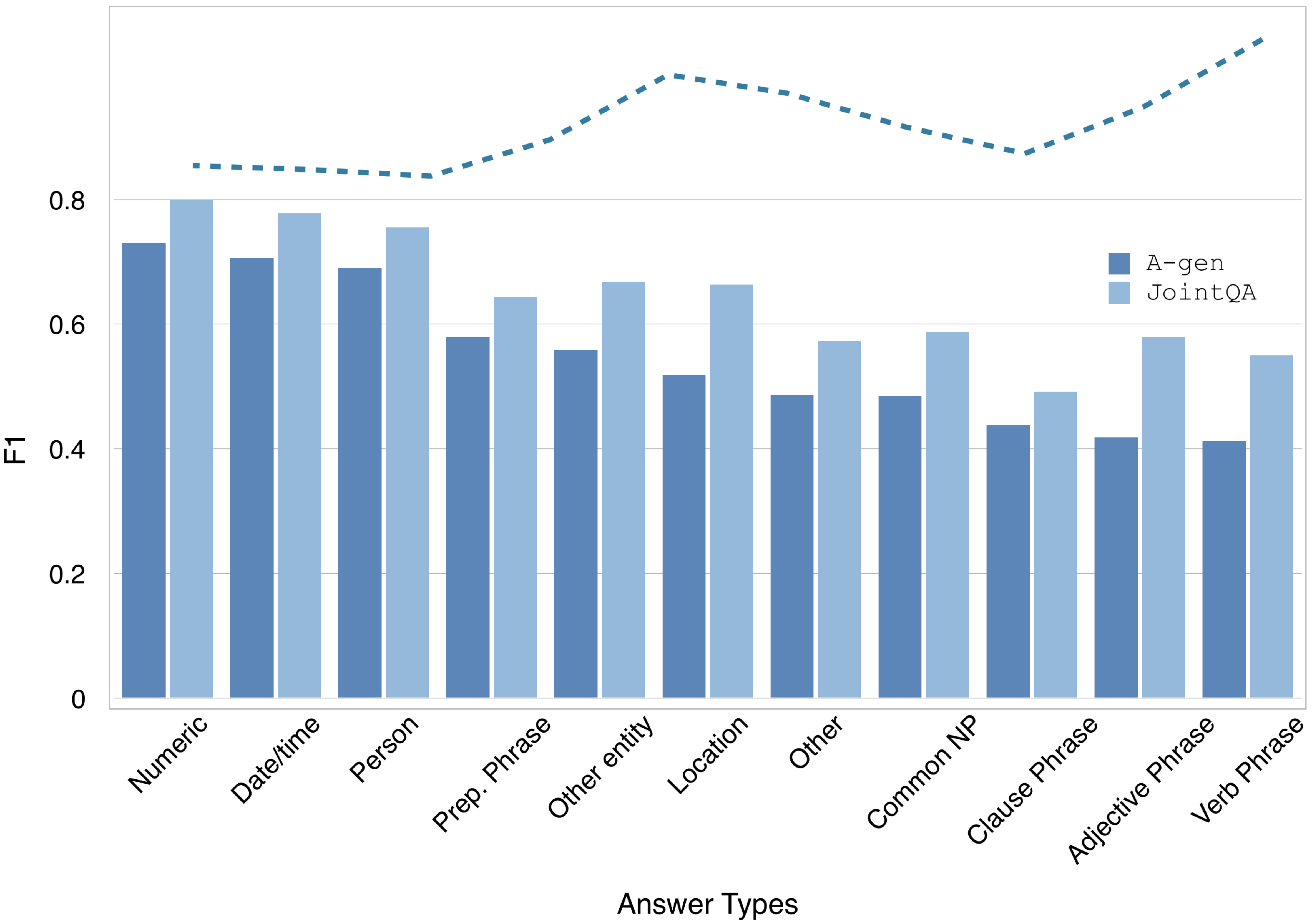}
    \caption{Comparison between \texttt{A-gen} and \texttt{JointQA} stratified by answer types. The dashed curve indicates period-2 moving average of the performance difference between the models.}
    \label{fig:at_strat}
\end{figure}

For a finer grained analysis, we first categorize test set answers based on their entity types, then stratify the QA performance comparison between \texttt{A-gen} and \texttt{JointQA}. The categorization relies on \textit{Stanford CoreNLP} \citep{manning2014stanford} to generate constituency parses, POS tags, and NER tags for answer spans (see~\citealt{rajpurkar2016squad} for more details). As seen in Figure~\ref{fig:at_strat}, the joint model significantly outperforms the single model in all categories. Interestingly, the moving average of the performance gap (dashed curve above bars) exhibits an upward trend as the \texttt{A-gen} model performance decreases across answer types, suggesting that the joint model helps most where the single model performance is weakest.

\subsection{Qualitative Examples}
\begin{table*}[t!]
  \caption{Examples of QA behaviour changes possibly induced by joint training. Gold answers correspond to text spans in green. In both the positive and the negative cases, the answers produced by the joint model are highly related (and thus presumably influenced) by the generated questions.}
  \centering
  \label{tab:cherrypick}
    \begin{tabular}{r|rp{0.68\textwidth}}
      \toprule
      Positive & Document & \textit{in the 1960 election to choose his successor , eisenhower endorsed his own vice president , republican {\color{green} richard nixon} against democrat {\color{red} john f. kennedy} .} \\
      & $\mathrm{Q_{gold}}$ & \textit{who did eisenhower endorse for president in 1960 ?} \\
      & $\mathrm{Q_{gen}}$ & \textit{what was the name of eisenhower 's own vice president ?} \\
      & Answer & {\tt A-gen}: \textit{\color{red} john f. kennedy}\hspace{25mm}{\tt JointQA:} \textit{\color{green} richard nixon} \\
      \midrule
      Negative & Document & \textit{in 1870 , tesla moved to karlovac , {\color{green} to attend school} at the higher real gymnasium , where {\color{red} he was profoundly influenced by a math teacher martin sekuli\'{c}}} \\
      & $\mathrm{Q_{gold}}$ & \textit{why did tesla go to karlovac ?} \\
      & $\mathrm{Q_{gen}}$ & \textit{what did tesla do at the higher real gymnasium ?} \\
      & Answer & {\tt A-gen}: \textit{{\color{green} to attend school} at the higher real gymnasium}\\
      & & {\tt JointQA:} \textit{\color{red} he was profoundly influenced by a math teacher martin sekuli\'{c}} \\
      \bottomrule
    \end{tabular}
\end{table*}

Qualitatively, we have observed interesting ``shifts'' in attention before and after joint training. For example, in the positive case in Table~\ref{tab:cherrypick}, the gold question asks about the direct object,\emph{Nixon}, of the verb \emph{endorse}, but the \texttt{A-gen} model predicts the indirect object, \emph{Kennedy}, instead. In contrast, the joint model asks about the appositive of \emph{vice president} during question generation, which presumably ``primes'' the model attention towards the correct answer \emph{Nixon}. Analogously in the negative example, QA attention in the joint model appears to be shifted by joint training towards an answer that is incorrect but closer to the generated question.

Note that the examples from Table~\ref{tab:cherrypick} come from the validation set, and it is thus not possible for the joint model to memorize the gold answers from question-generation mode --- the priming effect must come from some form of knowledge transfer between q-gen and a-gen via joint training.

\subsection{Implementation Details}
\label{sec:impl-details}
Implementation details of the proposed model are as follows. The encoder vocabulary indexes all words in the dataset. The decoder vocabulary uses the top 100 words sorted by their frequency in the gold questions in the training data. This encourages the model to generate frequent words (e.g. \emph{wh}-words and function words) from the decoder vocabulary and copy less frequent ones (e.g., topical words and entities) from the document.

The word embedding matrix is initialized with the 300-dimensional \emph{GloVe} vectors
\citep{penningtonglove}. The dimensionality of the character representations is 32. The number of hidden units is 384 for both of the encoder/decoder RNN cells. Dropout is applied at a rate of 0.3 to all embedding layers as well as between the hidden states in the encoder/decoder RNNs across time steps.

We use \emph{adam} \citep{kingma2014adam} as the step rule for optimization with mini-batch size 32. The initial learning rate is $2e-4$, which is decayed at a rate of 0.5 when the validation loss increases for two consecutive epochs.

The model is implemented using \textit{Keras} \citep{chollet2015keras} with the \textit{Theano} \citep{al2016theano} backend.

\section{Conclusion}
We proposed a neural machine comprehension model that can jointly ask and answer questions given a document. We hypothesized that question answering can benefit from synergistic interaction between the two tasks through parameter sharing and joint training under this multitask setting.
Our proposed model adopts an attention-based sequence-to-sequence architecture that learns to dynamically switch between copying words from the document and generating words from a vocabulary. Experiments with the model confirm our hypothesis: the joint model outperforms its QA-only counterpart by a significant margin on the SQuAD dataset.

Although evaluation scores are still lower than the state-of-the-art results achieved by dedicated QA models, the proposed model nonetheless demonstrates the effectiveness of joint training between QA and question generation, and thus offers a novel perspective and a promising direction for advancing the study of QA.

\bibliography{ref}
\bibliographystyle{icml2017}

\end{document}